# Deep Domain Adaptive Object Detection: a Survey

Wanyi Li
*Institute of Automation Chinese Academy of Sciences*
Beijing, China
wanyi.li@ia.ac.cn

Fuyu Li
*Institute of Automation Chinese Academy of Sciences*
Beijing, China
lifuyu2019@ia.ac.cn

Yongkang Luo
*Institute of Automation Chinese Academy of Sciences*
Beijing, China
yongkang.luo@ia.ac.cn
(Corresponding author)

Peng Wang
*Institute of Automation Chinese Academy of Sciences*
Beijing, China
peng_wang@ia.ac.cn

Jia sun
*Institute of Automation Chinese Academy of Sciences*
Beijing, China
jia.sun@ia.ac.cn

***Abstract*—Deep learning (DL) based object detection has achieved great progress. These methods typically assume that large amount of labeled training data is available, and training and test data are drawn from an identical distribution. However, the two assumptions are not always hold in practice. Deep domain adaptive object detection (DDAOD) has emerged as a new learning paradigm to address the above mentioned challenges. This paper aims to review the state-of-the-art progress on deep domain adaptive object detection approaches. Firstly, we introduce briefly the basic concepts of deep domain adaptation. Secondly, the deep domain adaptive detectors are classified into five categories and detailed descriptions of representative methods in each category are provided. Finally, insights for future research trend are presented.**

***Keywords—object detection, deep domain adaptation, and adaptive object detection***

## I. INTRODUCTION

Object detection is a fundamental and challenging task in computer vision and supports many applications such autonomous driving, robot vision and human computer interaction. State-of-the-art DL based object detection methods usually assume training data and test data are both drawn from an identical distribution. These detection models rely on seriously large amount of annotated training samples. In practice, collecting annotated data is expensive and sometime is not possible. Deep domain adaptation has emerged as a new learning paradigm to address the above mentioned challenges.

Following the success in some computer vision tasks including image classification and semantic segmentation brought by the deep domain adaptation (DDA) [1, 2], it is expected that utilizing DDA will improve the performance of object detection. Recent years, intensive research to experiment with DDA in the task of object detection is conducting and some DDAOD methods have been proposed.

There have been some works concentrating on review domain adaptation [1, 2] and deep learning based object detection [3], while there is still lack of a survey on DDAOD. The goal of this paper is to review the state-of-the-art progress on DDAOD methods and provide some insight for future research trend.

Funded by National Natural Science Foundation of China (Nos. 61771471, 91748131) and Beijing Municipal Natural Science Foundation (No. 4204113).

## II. DEEP DOMAIN ADAPTIVE OBJECT DETECTION

We use the narrow sense definition of deep domain adaptation (DDA) in [1]. DDA is based on deep learning architectures designed for DA and can obtain a firsthand effect from deep networks via back-propagation [1]. Deep domain adaptive object detection (DDAOD) aims to learn a robust object detector using label-rich data of source domain and label-agnostic or label-poor data of target domain, the learning process relies on the DDA models or principle used in DDA. The distributions between source domains and target domains are dissimilar or totally different. Trained object detector is expected to perform well in target domain.

In this section, we first introduce several factors that will be used later for categorization of DDAOD methods and then review related DDAOD methods. The categorization factors are as follows.

- Mechanism to address domain shift

There are five types of mechanism to address domain shift: discrepancy-based, adversarial-based, reconstruction-based, hybrid and others.

- One-step vs. multi-step adaptation methods

When source and target domains are directly related, transferring knowledge can be accomplished in one step. While there is little overlap between the two domains, a series of intermediate bridges are used to connect two seemingly unrelated domains and then perform one-step DA via this bridge, named multi-step DA.

- Labeled data of the target domain

In consideration of labeled data of the target domain, we can categorize DDAOD into supervised, semi-supervised, weakly-supervised, few-shot and unsupervised.

- Base detector

Domain adaptive detection approaches are usually based on some existed excellent detectors, such as Faster RCNN, YOLO, SSD, etc.

- Open source or not

This factor indicates whether the source code of method is available on line. If it is open source, the link will be provided.

According to the above listed categorization factors, we first classify DDAOD methods in TABLE I. , and then review them in the following subsections.

### A. *Discrepancy-based DDAOD*

Discrepancy-based DDAOD approaches diminish the domain shift by fine-tuning the deep network based detection model with labeled or unlabeled target data.

Khodabandeh et.al. [4] proposed a robust learning approach for domain adaptive object detection. The authors formulated the problem as training with noisy labels. Based on a set of noisy object bounding boxes obtained via a detection model trained only in the source domain, final detection model is trained.

To address domain shift from synthetic images to real images, Cai et.al. [5] advanced Mean Teacher paradigm to be applicable for cross domain detection and presented Mean Teacher with Object Relations (MTOR). This method novelly remolds Mean Teacher under the backbone of Faster R-CNN by integrating the object relations into the measure of consistency cost.

Cao et.al. [6] presented an auto-annotation framework to label pedestrian instance iteratively in visible and thermal channels, which leverages the complementary information in multispectral data. The auto-annotation framework consists of iterative annotation, temporal tracking and label fusion. To learn the multispectral features for robust pedestrian detection, the obtained annotations are then fed to a two-stream region proposal network (TS-RPN).

### B. *Adversarial-based DDAOD*

Adversarial-based DDAOD methods utilize domain discriminators and conduct adversarial training to encourage domain confusion between the source domain and the target domain. Domain discriminators classify whether a data point is drawn form the source or target domain.

Domain adaptive Faster RCNN [7] is the first work to deal with the domain adaptation problem for object detection. The authors used H-divergence to measure the divergence between data distribution of source domain and target domain, and conducted adversarial training on features. Three adaptation components are designed, i.e., image-level adaptation, instance-level adaptation and consistency check.

Motivated by the local nature of detection, Zhu et.al. [8] proposed a region-level adaptation framework. To address the questions of "where to look" and "how to align", effectively and robustly, two key components, region mining and adjusted region-level alignment, are designed. The second component aligns discriminative regions adversarially using two generators and two discriminators.

Wang et.al [9] proposed a few-shot adaptive Faster-RCNN framework, termed FAFRCNN. It consists of two level of adaptation modules, i.e., image-level and instance-level, coupled with a feature pairing mechanism followed domain classifier and a strong regularization for stable adaptation.

Saito et.al. [10] proposed an unsupervised adaptation method for object detection that combines weak global alignment with strong local alignment, called the Strong-Weak Domain Alignment model. The strong-local alignment is performed by a local domain classifier network and the weak-global alignment by a global domain classifier.

He et.al. [11] proposed a multi-adversarial Faster RCNN (MAF) detector for addressing unrestricted object detection problem. The method includes two modules, i.e., hierarchical domain feature alignment and aggregated proposal feature alignment.

Shen et.al [12] proposed a gradient detach based stacked complementary losses (SCL) method for unsupervised domain adaptive object detection. This method used multiple complementary losses for better optimization, and proposed gradient detach training to learn more discriminative representations.

Zhang et.al [13] proposed a synthetic-to-real domain adaptation method for object instance segmentation. There are three different feature adaptation modules, i.e., global-level base feature adaptation module, local-level instance feature adaptation module, and subtle-level mask feature adaptation module.

Zhuang et.al. [14] proposed Image-Instance Full Alignment Networks (iFAN) to tackle unsupervised domain adaptive object detection. There are two alignment modules: image-level alignment aligns multi-scale features by training adversarial domain in a hierarchically-nested fashion, and full alignment exploits deep semantic information and elaborate instance representations to establish a strong relationship among categories and domains.

To harmonize transferability and discriminability for adaptive object detectors, Chen et.al. [15] propose a Hierarchical Transferability Calibration Network (HTCN), which hierarchically (local-region/image/instance) calibrates the transferability of feature representations. Alignments in different level are achieved via adversarial training process and three domain discriminators are included in the structure of HTCN.

### C. *Reconstruction-based DDAOD*

Reconstruction-based DDAOD presume that the reconstruction of the source or target samples is helpful to improve the performance of domain adaptation object detection.

Arruda et.al. [16] proposed a cross-domain car detection method using unsupervised image-to-image translation. CycleGAN was explored to enable the generation of an artificial dataset (fake dataset) by translating images from day-time domain to night-time domain. Final detection model is trained on the fake dataset with the annotations transferred from source domain.

Lin et.al [17] introduced a multi-modal structure-consistent image-to-image translation model to realize domain adaptive vehicle detection. The image translation model generates diverse and structure-preserved translated images across complex domains.

Guo et.al. [18] presented an approach to pedestrian detection in thermal infrared images with limited annotations. To tackle the domain shift between thermal and color images, the authors proposed to learn a pair of image transformers to convert images between the two modalities, jointly with a pedestrian detector.

Devaguptapu et.al. [19] proposed to utilize image-to-image translation frameworks to generate pseudo-RGB equivalents of a given thermal image, and then to employ a multi-modal object detection architecture for thermal image.

Liu et.al. [20] presented an unsupervised image translation framework from thermal to visible, which is based on generative adversarial networks (GANs). The infrared-to-visible algorithm is referred as IR2VI. Object detector is trained on annotated visible images and applied directly to translated fake visible images.

### *D. Hybrid DDAOD*

Hybrid DDAOD use two or more aforementioned mechanisms simultaneously to obtain better performance.

Inoue et.al. [21] proposed the novel task, cross-domain weakly supervised object detection, in which image-level annotation is available in target domain. To address this task, a two-step progressive domain adaptation technique is proposed. This method fine-tunes the detector on two types of artificially and automatically generated samples. A CycleGAN based image-image translation is used to artificially generate samples, while automatically generated samples are obtained by pseudo-labeling.

Shan et.al. [22] presented a pixel and feature level based domain adaptive object detector. The method consists of two modules, pixel-level domain adaptation (PDA) mainly based on CycleGAN, and feature-level domain adaptation (FDA) based on Faster RCNN. The two modules can be integrated together and trained in an end-to-end way.

To alleviate the imperfect translation problem of pixel-level adaptations, and the source-biased discriminability problem of feature-level adaptations simultaneously, Kim et.al. [23] introduced a domain adaptive representation learning paradigm for object detection. It consists of Domain Diversification (DD) stage and Multidomain-invariant Representation Learning (MRL) stage.

Kim et.al. [24] introduced a domain adaptive one-stage object detection method consisting of a weak self-training (WST) method and adversarial background score regularization (BSR). WST manage to diminish the adverse effects of inaccurate pseudo-labels, while BSR reduce the domain shift by extracting discriminative features for target backgrounds.

Rodriguez et.al. [25] proposed a two-step domain adaptive detector, which is based on low-level adaptation via style transfer and high-level adaptation via robust pseudo labelling.

Hsu et.al. [26] proposed a progressive domain adaptive object detector. By translating the source images to mimic the ones of the target images, an intermediate domain is constructed. To address the domain-shift problem, the authors adopted adversarial learning to align distributions at the feature level and applied a weighted task loss to deal with unbalance image quality of the intermediate domain.

Yu et.al. [27] proposed the Cross-Domain Semi-Supervised Learning (CDSSL) framework to overcome limitation of previously many adversarial methods. The limitation is that they cannot address the domain content distribution gap which is also important for object detectors. The CDSSL framework leverages high quality pseudo labels to learn from target domain directly and conduct fine-grained domain transfer to reduce the style gap. Besides, progressive-confidence-based label sharpening and imbalanced sampling strategy are also included. Comparing best prior work on mAP, 2.2% - 9.5% performance gain was achieved.

Zheng et.al. [28] presented a coarse-to-fine feature adaptation approach for cross-domain two-stage object detection. It consists two adaptation modules, i.e., Attention-based Region Transfer (ART) and Prototype-based Semantic Alignment (PSA). ART extracts foreground regions and adopts attention mechanism, and aligns their feature distributions via multi-layer adversarial learning. PSA utilizes prototypes to perform conditional distribution alignment of foregrounds at the semantic level. According to conducted experiments, state-of-the-art results are reached.

### *E. Other DDAOD*

Other DDAOD methods can not be categorized into the four above mentioned categories. They use other mechanism such as graph-induced prototype alignment [29], categorical regularization [30] to seek for domain alignment.

To deal with the problems including align source and target domain on local instance level and class-imbalance in cross-domain detection tasks, Xu et.al. [29] propose the Graph-induced Prototype Alignment (GPA) framework and embed it into a two-stage detector, Faster R-CNN. Experimental results shown that the GPA framework outperforms existing methods with a large margin.

Considering previous work still overlook to match crucial image regions and important instances across domains, Xu et.al. [30] propose a categorical regularization framework. It can be utilized as a plug-and-play component on a number of Domain Adaptive Faster R-CNN methods. Two regularization modules are designed. The first module exploits the weakly localization ability of classification CNNs, while the second exploits the categorical consistency between image-level and instance-level predictions.

TABLE I. SUMMARY OF DEEP DOMAIN ADAPTIVE OBJECT DETECTION (DDAOD) METHODS

| **No** | **Method Year** | **One/multi-step DA** | **Labeled target** | **Basic detector** | **Open** | **Dataset [1]& task** | **mAP[2] (%)** |
|---|---|---|---|---|---|---|---|
| | | | | **Discrepancy-based** | | | |
| 1 | Khodabandeh et.al. [4], 2019 | One-step DA | Unsupervised | Faster RCNN | A link is provided but without code yet.<br>https://github.com/mkhodabandeh/robust_domain_adaptation | Cityscapes [31] → Foggy[32]<br>Cityscapes → KITTI [33]<br>KITTI → Cityscapes<br>SIM 10k [34] → Cityscapes | 36.5, oracle: 43.5<br>**car AP: 77.6**, oracle: 90.1<br>car AP: 43.0, oracle: 68.1<br>car AP: 42.6, oracle[3]: 68.1 |
| 2 | Cai et.al. [5], 2019 | One-step DA | Unsupervised | Faster RCNN | No | Cityscapes → Foggy<br>SIM 10k → Cityscapes<br>Synthetic [35]→COCO [36]<br>Synthetic→YTBB [37] | 35.1<br>car AP: 46.6<br>20.7<br>22.8 |
| 3 | Cao et.al. [6] ,2019 | One-step DA | Unsupervised | Faster RCNN Replaceable | No | Caltech [38] visible→KAIST multispectral [39] | F1 of annotation: 0.75<br>Miss rate:32.66 |
| **Adversarial-based** | | | | | | | |
| 4 | Chen et.al. [7], 2018 | One-step DA | Unsupervised | Faster RCNN | Yes, https://github.com/yuhuayc/da-faster-rcnn | Cityscapes → Foggy<br>Cityscapes → KITTI<br>KITTI → Cityscapes<br>SIM 10k → Cityscapes | 27.6<br>car AP : 64.1<br>car AP : 38.5<br>car AP: 39.0 |
| 5 | Zhu et.al. [8], 2019 | One-step DA | Unsupervised | Faster R-CNN | Yes, https://github.com/xinge008/SCDA | Cityscapes → Foggy<br>KITTI → Cityscapes<br>SIM 10k → Cityscapes | 33.8<br>car AP: 42.5<br>car AP: 43.0 |
| 6 | Wang et.al. [9][4], 2019 | One-step DA | Few-shot | Faster R-CNN | A link is provided but without code yet. https://github.com/twangnh/FAFRCNN | Cityscapes → Foggy<br>Cityscapes→Udacity<br>SIM 10k → Cityscapes<br>SIM10K→Udacity<br>Udacity [40]→ Cityscapes | 31.3<br>48.5<br>car AP: 41.2<br>car AP: 40.5<br>50.2 |
| 7 | Saito et.al. [10], 2019 | One-step DA | Unsupervised | Faster R-CNN | Yes, https://github.com/VisionLearningGroup/DA_Detection | Cityscapes→ Foggy<br>PASCAL [41]→Clipart [21]<br>PASCAL→Watercolor [21]<br>SIM 10k → Cityscapes | 34.3<br>38.1<br>**53.3**<br>**car AP:47.7** |
| 8 | He et.al. [11], 2019 | One-step DA | Unsupervised | Faster R-CNN | No | Cityscapes→ Foggy<br>Cityscapes → KITTI<br>KITTI → Cityscapes<br>SIM 10k → Cityscapes | 34.0<br>car AP:72.1<br>car AP :41.0<br>car AP: 41.1 |
| 9 | Shen et.al [12], 2019 | One-step DA | Unsupervised | Faster R-CNN | Yes, https://github.com/harsh-99/SCL | Cityscapes→ Foggy<br>Cityscapes→ KITTI<br>KITTI → Cityscapes<br>PASCAL→Clipart<br>PASCAL→Watercolor<br>SIM 10k → Cityscapes | 37.9<br>**car AP:72.7**<br>41.9<br>41.5<br>**55.2**<br>car AP: 42.6 |
| 10 | Zhang et.al [13], 2019 | One-step DA | Unsupervised | Mask R-CNN | No | SYNTHIA [42]→ Cityscapes<br>VKITTI [43]→ Cityscapes | 33.2<br>car AP: 52.8 |
| 11 | Zhuang et.al. [14], 2020 | One-step DA | Unsupervised | Faster R-CNN | No | Cityscapes→ Foggy<br>SIM 10k → Cityscapes | 36.2<br>car AP:47.1 |
| 12 | Chen et.al. [15], 2020 | One-step DA | Unsupervised | Faster R-CNN | Yes https://github.com/chaoqichen/HTCN | Cityscapes→ Foggy<br>PASCAL → Clipart1k<br>Sim10k → Cityscapes | 39.8, oracle: 40.3<br>40.3<br>42.5 |
| **Reconstruction-based** | | | | | | | |
| 13 | Arruda et.al. [16], 2019 | One-step DA | Unsupervised | Faster R-CNN, Replaceable | Yes, https://github.com/123zhen123/publications-arruda-ijcnn-2019 | BDD100k [44], day →night | **86.6±0.7**, oracle: 92.0±0.8 |
| 14 | Lin et.al [17],2019 | One-step DA | Unsupervised | YOLO / Faster R-CNN | No | BDD100k, day → night | mAP of YOLO / Faster R-CNN: **41.9/67.0** |
| 15 | Guo et.al. [18], 2019 | One-step DA | Supervised | Faster RCNN, replaceable | No | KAIST visible→ thermal | Miss rate: 42.65 |
| 16 | Devaguptapu et.al. [19],2019 | One-step DA | Supervised | Faster RCNN | Yes, https://github.com/tdchaitanya/MMTOD | FLIR ADAS [45], visible →thermal<br>KAIST visible→ thermal | 61.54<br>53.56 |
| 17 | Liu et.al. [20],2019 | One-step DA | Unsupervised | Faster RCNN, replaceable | No | SENSIAC [46], visual → middle-wave infrared | 91.7 |
| Hybrid | | | | | | | |
| 18 | Inoue et.al. [21], 2018 | Multi-step DA | Weakly-Supervised | SSD, replaceable | Yes. https://naoto0804.github.io/cross_domain_detection/ | PASCAL → Clipart1k<br>→ Watercolor2k<br>→ Comic2k | **46.0**, Ideal case: 55.4<br>**54.3**, Ideal case: 58.4<br>**37.2**, Ideal case: 46.4 |
| 19 | Shan et.al. [22], 2019 | Multi-step DA | Unsupervised | Faster RCNN | No | Cityscapes → Foggy<br>Cityscapes → KITTI<br>KITTI → Cityscapes<br>KITTI→VKITTI-Rainy<br>Sim10k → Cityscapes<br>Sim10k → KITTI | mAP: 28.9<br>car AP: 65.6<br>car AP: 41.8<br>mAP: 52.2<br>car AP: 39.6<br>car AP: 59.3 |
| 20 | Kim et.al. [23], 2019 | Multi-step DA | Unsupervised | Faster RCNN | Yes, https://github.com/TKKim93/DivMatch | Cityscapes → Foggy<br>PASCAL → Clipart1k<br>→Watercolor2k<br>→ Comic2k | 34.6<br>**41.8**<br>52.0<br>**34.5** |
| 21 | Kim et.al. [24], 2019 | One-step DA | Unsupervised | SSD | No | PASCAL → Clipart1k<br>→Watercolor2k<br>→Comic2k | 35.7<br>49.9<br>26.8 |
| 22 | Rodriguez et.al. [25], 2019 | Multi-step DA | Unsupervised | SSD | No | Cityscapes → Foggy<br>PASCAL → Clipart1k<br>→Watercolor2k,Comic2k<br>Sim10k → Cityscapes | 29.7<br>**44.8**<br>**57.3, 39.4**<br>car AP:44.2 |
| 23 | Hsu et.al. [26], 2020 | Multi-step DA | Unsupervised | Faster RCNN | Yes<br>https://github.com/kevinhkhsu/DA_detection | Cityscapes→BDD100k<br>Cityscapes → Foggy<br>KITTI → Cityscapes | 24.3, oracle: 43.3<br>**36.9**, oracle: 39.2<br>**car AP: 43.9** oracle: 55.8 |
| 24 | Yu et.al. [27], 2019 | Multi-step DA | Unsupervised | Faster RCNN | A link is provided but without code yet.<br>https://github.com/Mrxiaoyuer/CDSSL | Cityscapes → Foggy<br>KITTI → Cityscapes<br>Sim10k → Cityscapes | **38.2**,oracle:42.5<br>**Car AP:46.4**, oracle:62.7<br>**car AP:52.3**, oracle: 62.7 |
| 25 | Zheng et.al. [28] . 2020 | One-step DA | Unsupervised | Faster RCNN | No | Cityscapes → Foggy<br>Cityscapes → KITTI<br><br>Sim10k → Cityscapes | **38.6**, oracle:43.3<br>**car AP: 73.6**, oracle:88.4<br>(AP: 41.0, oracle:85.4)<br>car AP: 43.8, oracle:59.9 |
| **Others** | | | | | | | |
| 26 | Xu et.al. [29], 2020 | One-step DA | Unsupervised | Faster RCNN | Yes<br>https://github.com/ChrisAllenMing/GPA-detection | Cityscapes → Foggy<br>KITTI → Cityscapes<br>Sim10k → Cityscapes | **39.5**<br>**Car AP: 47.9**<br>**Car AP: 47.6** |
| 27 | Xu et.al, [30], 2020 | One-step DA | Unsupervised | Faster RCNN | Yes<br>https://github.com/Megvii-Nanjing/CR-DA-DET | Cityscapes→BDD100k<br>Cityscapes → Foggy<br>PASCAL → Clipart1k | 26.9, oracle:38.6<br>37.4, oracle:42.4<br>38.3 |

Note: 1.The source of dataset can be found in the original reference. 2 Bolded red, green and blue highlights the first place, second place and third place respectively.3. Oracle represents Faster R-CNN trained on target domain. 4. Results of UDA are included in this table.

## III. Conclusion and Future Directions

This paper surveyed 27 Deep domain adaptive object detection (DDAOD) approaches. The reviewed methods are summarized and categorized according to our presented five category factors. Performance on different domain adaptive object detection tasks is also compared. It is seen that hybrid methods achieved most top places, adversarial-based gets more and the others get least top places. It is shown that adversarial training and incorporating more adaptation mechanisms works better. Although various DDAOD methods have been proposed in recent years, a clear margin still exists between achieved performance and the oracle results of detector trained on labelled target data. Thus there are much works to conduct further. Some of them are stated as follows.

A promising solution is to further combine the merits of different category adaptation methods, like [25], which combines style transfer and robust pseudo labelling and obtains better performance. A possible combination is training a detector adversarially and using trained detector to generate pseudo labels for target samples.

Another promising direction is to explore the local nature of detection. For example, generating simulated instance-level samples that are similar to instance-level samples of target domain, and then synthesizing training samples for detection using generated instance-level image patches and background images of target domain.

Thirdly, most reviewed works deal with homogeneous DDAOD, while heterogeneous DDAOD is more challenge since there is larger domain gap. Thus, it worth conducting more research such as adaptation from visible domain with large amount of labeled data to thermal infrared domain for which annotated data is expensive to collect. Works with high impact in this direction are expected.

Finally, utilizing state-of-the-art domain adaptive classification models and embedding into detection framework, and exploring the domain-shift detection from scratch is also a promising direction.